# We Need Knowledge Distillation for Solving Math Word Problems

Zhenquan Shen[a], Xinguo Yu[a], Xiaotian Cheng[a], Rao Peng[a], Hao Ming[a]

[a] *Faculty of Artificial Intelligence in Education, Central China Normal University, Wuhan 430079, China*

**Abstract**

The enhancement of mathematical capabilities in large language models (LLMs) fosters new developments in mathematics education within primary and secondary schools, particularly as they relate to intelligent tutoring systems. However, LLMs require substantial computational resources, resulting in significant costs in educational contexts. To mitigate this drawback, this paper investigates the feasibility of compressing LLMs for solving math word problems (MWPs). We compress the embedded vectors encoded by BERT and distill a considerably smaller student model. Our findings indicate that the student model can maintain nearly 90% of the performance of the teacher model while utilizing only 1/12 of its parameters. In addition to achieving high accuracy, the model exhibits strong generalizability, as the compressed vectors perform well across all tasks related to MWPs, and the distillation process is not task-specific. The success of this distillation demonstrates that the underlying principles are generic and not limited to a specific task. We further explore the reasons behind the compressibility of embedded vectors, revealing that part-of-speech information, rather than entity recognition, is crucial for MWPs, which may significantly contribute to their compressibility. The improvements in efficiency and cost reduction provide substantial value for intelligent tutoring systems and significantly advance the field of intelligent education.

# 1. Introduction

In recent years, the rapid development of artificial intelligence has enabled more people to benefit from technological advancements [1]. Intelligent tutoring systems powered by AI are transforming traditional teaching methods in the field of education [2]. These systems not only provide real-time problem-solving support and personalized learning pathways, but also enhance learning feedback and simulate human tutoring functions. Together, these features contribute to improving students' learning efficiency. Given the importance of mathematics education, it is crucial to develop intelligent tutoring systems specifically for mathematics. Mathematical ability is a key indicator of artificial intelligence [3-5], and math solvers are the core technology behind math-focused intelligent tutoring systems [6-11].

There is a significant demand for math word problems (MWP) in primary and secondary education. While the development of large language models (LLMs) has opened up new opportunities for solving math word problems with greater accuracy [1, 2], it has also introduced challenges that must be addressed. LLMs require substantial computational resources due to their size during the inference stage in practical deployment [12-14], which can lead to high costs and hinder commercialization. Moreover, the high latency of LLMs may negatively affect the user experience, ultimately impacting students' learning outcomes [15, 16].

Is there a way to reduce resource consumption, lower costs, and speed up the inference process while maintaining the mathematical capabilities of large models? This paper proposes a solution through knowledge distillation, which leverages the capabilities of LLMs while eliminating redundant information to enhance efficiency. The word vectors encoded by the original BERT model are compressed, and the underlying mechanism behind this compression is further explored. The remaining sections of this paper are organized as follows: The Related Work section discusses existing approaches to solving MWPs and knowledge distillation. The Method section outlines the process of distilling a smaller model and the criteria used for model evaluation. The Results section presents the evaluation outcomes. The Discussion section examines the underlying mechanisms behind these results. Finally, the Conclusion section summarizes the entire paper.

**2. Related Work**

2.1 Solving Math Word Problems

Math word problems (MWPs) are a type of problem-solving exercise that involves describing a situation or scenario using natural language, requiring the application of mathematical concepts and reasoning to find a solution. These problems bridge the gap between theoretical mathematics and real-world applications, fostering critical thinking and analytical skills. AI-powered solutions for MWPs streamline the teaching process and enrich the learning experience. They provide immediate feedback, personalized guidance, and interactive engagement, while also offering educators valuable insights. By integrating AI into mathematics education, institutions can significantly enhance both teaching efficiency and learning outcomes.

There are two primary methods for solving MWPs. The first is the model-matching method, represented by syntax-semantics models [6, 17-19]. This approach relies on predefined grammatical rules to extract explicit mathematical relations from textual descriptions and mine implicit relations from a knowledge base. These extracted relations are then sent to symbolic solvers to formulate equations and obtain solutions. The second method is based on deep neural networks, typically comprising an encoder and a decoder, which form an end-to-end solution. He et al. [20] compared typical neural solvers for MWPs. The Deep Neural Solver was the first deep learning algorithm capable of translating problem texts into equations without relying on manually crafted rules [21]. MWP solvers are generally designed using an Encoder-Decoder framework [22]. The encoder is responsible for learning the semantic representation and logical relationships within the problem text, whether explicitly or implicitly. MathEN [23] explored various sequence models for this task. The decoder, typically structured as a sequence or tree model, treats the math equation as a symbolic sequence consisting of numbers and operators to decode [20]. Several tree-structured models, such as Tree-Dec [24] and GTS [25], have been designed and widely accepted for math equation decoding, enhancing the generation of mathematical equations.

Recently, pre-trained models like BERT [26] and GPT [27] have been incorporated into MWP solvers to better represent background knowledge. Liang et al. [28] utilized BERT and RoBERTa for contextual number representation. Decode-only pre-trained LLMs, such as GPT [27], PaLM [29, 30], and LLaMA [31], demonstrate strong reasoning abilities, showcasing their potential in solving MWPs, especially when integrated with prompt engineering [32] and chain-of-thought reasoning [33]. These models have led to significant improvements in answer accuracy. For example, GPT-4 achieved nearly a 20% increase in answer accuracy on the MATH dataset compared to GPT-3.5 [34]. Although numerous LLMs have been proposed, BERT-based models have been found to achieve the highest accuracy for MWPs, particularly on the Math23k dataset, according to He et al. [20]. However, LLMs require substantial computational resources during the inference stage due to their size, leading to significant operational costs and posing challenges to commercialization. Additionally, the high latency associated with LLMs can negatively impact the user experience, potentially hindering students' learning outcomes.

 2.2 Knowledge Distillation

Knowledge distillation aims to transfer the knowledge from a large teacher network to a smaller student network. The student network is trained to mimic the behavior of the teacher network [35]. The key concept is to transfer the knowledge of the teacher model, which has been trained on a large dataset and exhibits high performance, into the student model, which is typically faster, more efficient, but less powerful [36].

This is accomplished by using the teacher's output — rather than the ground truth labels — as soft targets during training. This approach allows the student model to learn not only the final predictions but also the underlying patterns and decision boundaries that the teacher model has learned, leading to a more effective and efficient student model [36-38].

Knowledge distillation is widely applied in scenarios where deploying the large teacher model is impractical due to memory, computational power, or latency constraints. For instance, in mobile or embedded devices, the student model can provide near real-time predictions while maintaining accuracy comparable to that of the teacher model [36-38].

Several studies have focused on distilling BERT. TinyBERT [35] proposes a two-stage learning framework that distills the teacher model during both pre-training and fine-tuning. This method resulted in a 4-layer BERT with a 7.5-fold reduction in parameters and a 9.4-fold speed improvement, achieving 96.8% of the teacher model's performance. Furthermore, a 6-layer model trained using this technique nearly matched the performance of BERT-base. Distilled BiLSTM [39] introduces a method where BERT-large is distilled into a single-layer BiLSTM, reducing the number of parameters by 100 times and increasing speed by 15 times. BERT-PKD [40], unlike previous approaches, introduces Patient Knowledge Distillation. This method extracts knowledge from the intermediate layers of the teacher model, preventing the student model from fitting too quickly when distilling only the final layer. MobileBERT [41] focuses on reducing the dimensionality of each layer. While maintaining 24 layers, it reduces the number of parameters by 4.3 times, increases speed by 5.5 times, and scores on average only 0.6 points lower than BERT-base on the GLUE benchmark. Previous models have largely exhausted the distillation potential within BERT, but MiniLM [42] proposes a new approach by distilling the Value-Value matrices. However, existing distillation techniques have not thoroughly analyzed the compressibility of encoding vectors or addressed specific challenges related to math problems.

## 3. Method

Given the excellent performance of pre-trained language models on MWP tasks, we use them as the teacher model for the encoder to transform text into vectors. Through knowledge distillation, we transfer their encoding capabilities to a smaller student model. The effectiveness of this approach is then tested across a series of math-related tasks, with each task being implemented through a decoder.

Considering BERT's outstanding performance on MWP tasks, we aim to fully leverage its natural language encoding abilities. As a result, BERT is selected as the teacher model for knowledge distillation.

The pipeline of the entire method is shown in Fig. 1. The words of the MWP are input into the original BERT model, which generates encoded vectors, denoted as Vector 1. These encoded vectors are then passed through a compressor, reducing their dimensionality to a specified size. The resulting vectors are referred to as Vector 2. At this point, the encoding process of the teacher model is complete.

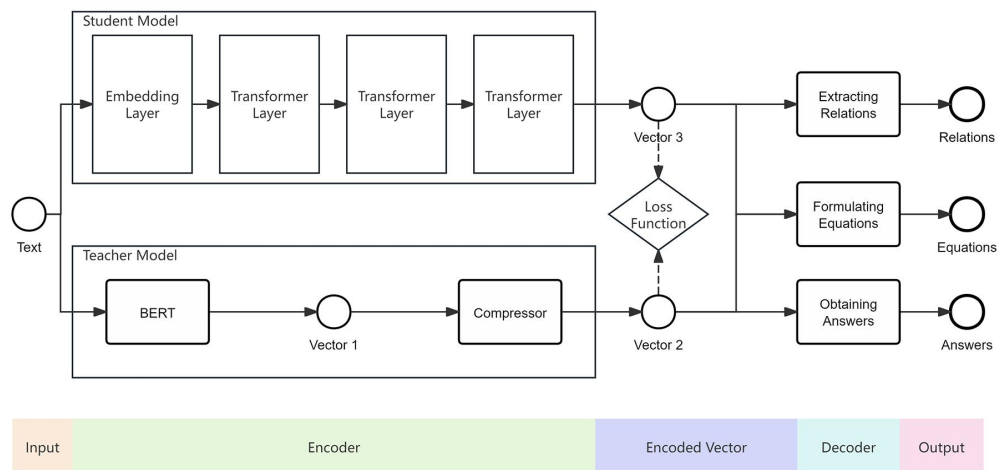

Fig.1 The overall method pipeline consists of a teacher model, which includes the original BERT and a compressor, and a student model composed of an embedding layer and three transformer layers. The distillation process is guided by a loss function based on the mean squared error of the encoded vectors from both the teacher and student models. To evaluate the performance of the compressed vectors, three distinct decoders related to MWP tasks are implemented, focusing on extracting relations, formulating equations, and obtaining answers, respectively.

To distill a student model capable of inheriting the encoding ability of the teacher model, the loss function for the knowledge distillation process is defined as follows:

$$\mathcal{L}_{KD} = MSE(z^T/t, z^S/t) \tag{1}$$

where $z^T$ and $z^S$ are the vectors encoded by teacher and student respectively, $MSE(\ )$ means the mean squared error loss function.

3.1 Student Model

The student model consists of an embedding layer followed by three transformer layers. Each transformer layer has a size of 256 with 16 attention heads.

A standard transformer layer includes two main sub-layers: multi-head attention (MHA) and a fully connected feed-forward network (FFN). The calculation of the attention function depends on three components: queries, keys, and values, which are represented as matrices $Q$, $K$, and $V$, respectively. The attention function can be formulated as follows:

$$A = \frac{QK^T}{\sqrt{d_k}} \tag{2}$$

$$Attention(Q, K, V) = softmax(A)V \tag{3}$$

where $d_k$ is the dimension of keys and acts as a scaling factor, $A$ is the attention matrix calculated from the compatibility of $Q$ and $K$ by dot-product operation. The final function output is calculated as a weighted sum of values $V$, and the weight is computed by applying $softmax()$ operation on the each column of matrix $A$. Multi-head attention is defined by concatenating the attention heads from different representation subspaces as follows:

$$MHA(Q,K,V) = Concat(h_1,\cdots,h_k)W \tag{4}$$

where $k$ is the number of attention heads, and $h_i$ denotes the $i$-th attention head, which is calculated by the $Attention()$ function with inputs from different representation subspaces. The matrix $W$ acts as a linear transformation.

Transformer layer also contains a fully connected feed-forward network, which is formulated as follows:

$$FFN(x) = Max(0, xW_1 + b_1)W_2 + b_2 \tag{5}$$

the FFN contains two linear transformations and one ReLU activation. For more details, readers can refer to [35].

3.2 Teacher Model

BERT-base is used as the teacher model, consisting of an embedding layer followed by 12 transformer layers. Each transformer layer has a size of 768 with 12 attention heads.

To minimize computational costs and enhance efficiency, the encoded vectors (Vector 1) generated by BERT-base are passed through a compressor for dimensionality reduction.

3.3 Compressor

There are many compressors that enable dimensionality reduction of the vectors. We selected the following types of compressors based on their applicability and efficiency considerations.

**Linear Layer.** This is the simplest compressor, where a linear layer is added directly after the last layer of the BERT to reduce the vectors to the specified dimensions.

**Principal Component Analysis (PCA).** PCA is a widely-used dimensionality reduction technique in data analysis and machine learning. PCA aims to reduce the dimensionality of a dataset while retaining as much information as possible. This is done by transforming the data into a new set of orthogonal axes called principal components. The first step is often to standardize the data. This involves subtracting the mean of each feature and scaling it to have unit variance. Standardization ensures that each feature contributes equally to the analysis, regardless of its original scale. PCA computes the covariance matrix of the data, which captures the relationships between different features. The covariance matrix helps identify the directions in which the data varies the most. PCA then calculates the eigenvalues and eigenvectors of the covariance matrix. The eigenvectors are the directions of maximum variance in the data, while the eigenvalues indicate the magnitude of this variance. The eigenvectors are sorted by their corresponding eigenvalues in descending order. The top eigenvectors form a new basis for the data. By projecting the original data onto these principal components, its dimensionality is reduced while preserving most of the variability. To reduce dimensionality, the top k principal components are selected and project the data onto this lower-dimensional space. This results in a new representation of the data with reduced dimensions but preserved structure. PCA is commonly used for visualizing high-dimensional data, noise reduction, and feature extraction, making it a valuable tool for exploratory data analysis and preprocessing in machine learning. Details of this method can be found in [43].

**Locally Linear Embedding (LLE).** LLE is a dimensionality reduction technique used in machine learning and data analysis. It helps to map high-dimensional data into a lower-dimensional space while preserving local structures. LLE starts by identifying local neighborhoods around each data point. For each point, it looks at its nearest neighbors in the high-dimensional space. For each point, LLE computes the weights that best reconstruct the point as a linear combination of its neighbors. These weights are chosen to minimize the reconstruction error, which is the difference between the point and its weighted sum of neighbors. Once the reconstruction weights are determined, LLE seeks a lower-dimensional representation of the data where the same local linear relationships are maintained. This involves solving an optimization problem to find the lower-dimensional coordinates that best preserve the relationships defined by the weights. Although LLE focuses on preserving local relationships, it often results in a meaningful global structure in the lower-dimensional space. LLE is particularly useful for visualizing high-dimensional data and uncovering underlying patterns by reducing the dimensionality while keeping the data's intrinsic geometric properties intact. Details of this method can be found in [44].

**Multidimensional Scaling (MDS).** MDS is a technique used for visualizing the similarity or dissimilarity between data points in a lower-dimensional space. It's particularly useful for exploring relationships within data and understanding complex datasets. The goal of MDS is to find a lower-dimensional representation of data such that the distances between points in this lower-dimensional space approximate the distances between points in the original high-dimensional space as closely as possible. MDS starts with a distance matrix that captures the pairwise distances between all pairs of data points. These distances can be computed using various metrics, such as Euclidean distance or other distance measures. MDS then seeks a configuration of points in a lower-dimensional space that preserves these distances as well as possible. This involves solving an optimization problem to minimize the difference between the distances in the lower-dimensional space and the original distances. The quality of the configuration is often evaluated using a cost function, which measures how well the distances in the lower-dimensional representation match the original distances. The goal is to minimize this cost function during the optimization process. MDS is particularly useful for visualizing high-dimensional data in a way that captures the relationships between data points, making it easier to interpret and analyze complex datasets. Details of this method can be found in [45].

**t-distributed Stochastic Neighbor Embedding (t-SNE).** t-SNE is a popular non-linear dimensionality reduction technique, primarily used for visualizing high-dimensional data in a low-dimensional space (usually 2D or 3D). It is widely applied in fields like machine learning, data mining, and computer vision for exploring data structures and patterns[46].

**Isometric Mapping (ISOMAP). ISOMAP** is a non-linear dimensionality reduction algorithm that extends classical multidimensional scaling (MDS) to better preserve the intrinsic geometric structure of high-dimensional data. It is particularly effective for data lying on a non-linear manifold and is widely used in tasks involving high-dimensional visualization, pattern recognition, and data analysis[47].

3.4 Decoder

A good encoder should be capable of performing well on a sufficient number of downstream tasks related to MWPs. In this context, decoders are used to evaluate the performance of encoders, which include both the teacher and student models.

For the first time, we propose that the model's ability to solve MWPs should be evaluated from three aspects: extracting mathematical relations, formulating equations, and obtaining answers. These three aspects — relation, equation, and answer — correspond to the sequential steps the human brain follows when solving MWPs.

A neural network is employed to extract mathematical relations especially implicit relations. a neural network miner based on quantity to relation attention neural network (QRAN) to mine implicit relations. The procedure consists of three steps: The first step is to encode the given problem into a sequence of vectors. A given problem can be tokenized as $\mathcal{P} = \{\omega_i\}_{i=1}^n$, each token $\omega_i$ can be represented as a word-context feature vector $v_i$ by encoders. Thus, $\mathcal{P}$ can be denoted by a sequence of vectors as $V = \{v_i\}_{i=1}^n$. The next process is to select the vectors related to quantity, including the numeric words like "100", "1/2" and the descriptive words like "double", "half". Let $N$ denote the set of the quantity vectors in the problem, and place $v_i$ into

$N$ if $\omega_i$ is a quantity word. Thus, $N = \{q_i\}_{i=1}^{k}$ contains all the quantity vectors, where $k$ is the total number of the quantity words in a problem. The second step is to obtain the goal vector $v_g$ representing implicit relations by adopting the quantity-relation attention mechanism. The concrete computing process is as follows:

$$\mu_i = \alpha \cdot \tanh(W_r \cdot [\bar{v}, q_i]) \; for \; i = 1, 2, \cdots, k \tag{6}$$

$$a_i = \frac{exp(\mu_i)}{\sum_{j-1}^{k} exp(\mu_j)} \tag{7}$$

$$v_g = \sum_{i=1}^{k} a_i \cdot q_i \tag{8}$$

where $\bar{v}$ is the average vector of the sequence $V$, $\mu_i$ is the relevance score between the whole problem and a quantity related word, $a_i$ is the attention score of each quantity in a softmax manner, $\alpha$ and $W_r$ are parameters trained for each specific implicit mathematical relation.

The goal vector $v_g$ can be transformed to an indicator $\hat{y}$ through a dense layer, judging whether an implicit relation needs to be added. The $\hat{y}$ is defined as follow:

$$\hat{y} = \sigma(W_c \cdot v_g + \beta_c) \tag{9}$$

where $\sigma$ is the sigmoid function, $W_c$ and $\beta_c$ are trainable parameters. For more details, readers can refer to [6].

A GTS decoder is employed to formulate equations and obtaining answers, details of the method can be found in [28]. By taking the problem embedding Vector 2 or Vector 3, the decoder generates a binary tree in a top-down manner for the solution, as tree-based representations have been commonly used in MWP solving. It is easy to encode operation order into the top-down structure of a tree, avoiding brackets and simplifying the equation representation by tree traversal.

## 4. Results

The publicly available MWP dataset, Math23K, was used to test the methods described above. All experiments in this study were conducted on an A800 GPU.

4.1 Compressibility of Encoded Vectors

We use the compressors mentioned above to reduce the dimensionality of the word vectors encoded by BERT. The linear layer compressor, due to its simplicity and low computational cost, is chosen as the primary test subject. The other compressors are used for supplementary validation to demonstrate the generalizability of the conclusions.

We extract the vector representations from both BERT and fine-tuned BERT on Math23K and analyze their characteristics to determine the optimal method for vector reduction. The results are shown in Fig. 2.

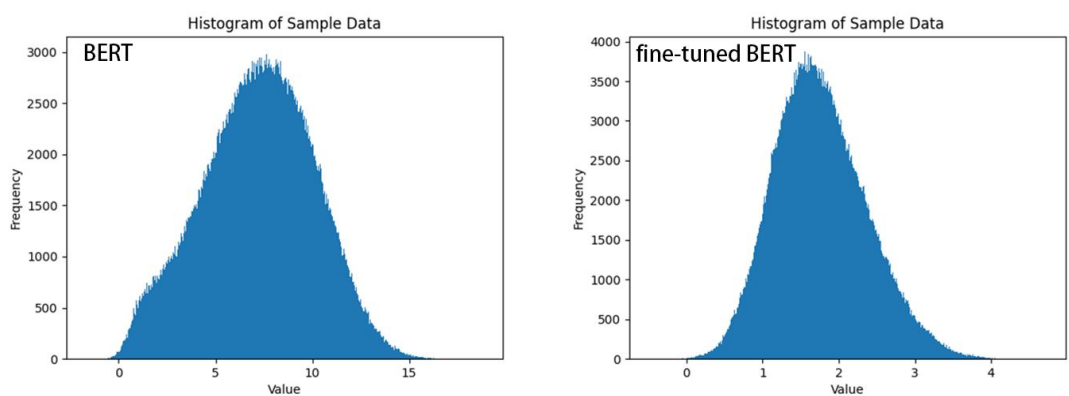

Fig.2 The distribution of BERT and fine-tuned BERT.

The vectors extracted from BERT approximately follow a normal distribution, and the vectors from fine-tuned BERT retain this property. This phenomenon suggests that the vector representations from large language models (LLMs) tend to cluster around middle values across different dimensions, indicating redundancy in these dimensions. To address this, we sort the dimensions of the sentence vectors based on their mean and variance values and then select the top 300 dimensions from the original vector as input for the decoder. However, the answer accuracy on Math23K is 12%, which is lower than expected. This indicates that pruning the dimensions of the BERT vector to achieve dimensionality reduction is ineffective. Thus, pruning the vector representation is not suitable for obtaining smaller vectors intended for math problem understanding. The reduction of vector dimensions must be adapted to the specific downstream tasks.

Other dimensionality reduction methods, such as PCA, KPCA, MDS, and LLE, are also used to reduce the dimensions of BERT. Among these, PCA is a linear approach, while the others are nonlinear. The vector distributions on a 2D plane using different methods are shown in Fig. 3. The degree of self-similarity between the original and reduced vectors is listed in Table 1. Interestingly, the vector reduced through PCA exhibits the highest similarity in distribution to BERT. This demonstrates the linear properties of BERT's vector representations and supports the rationale behind our approach of adding a linear layer to obtain reduced vectors.

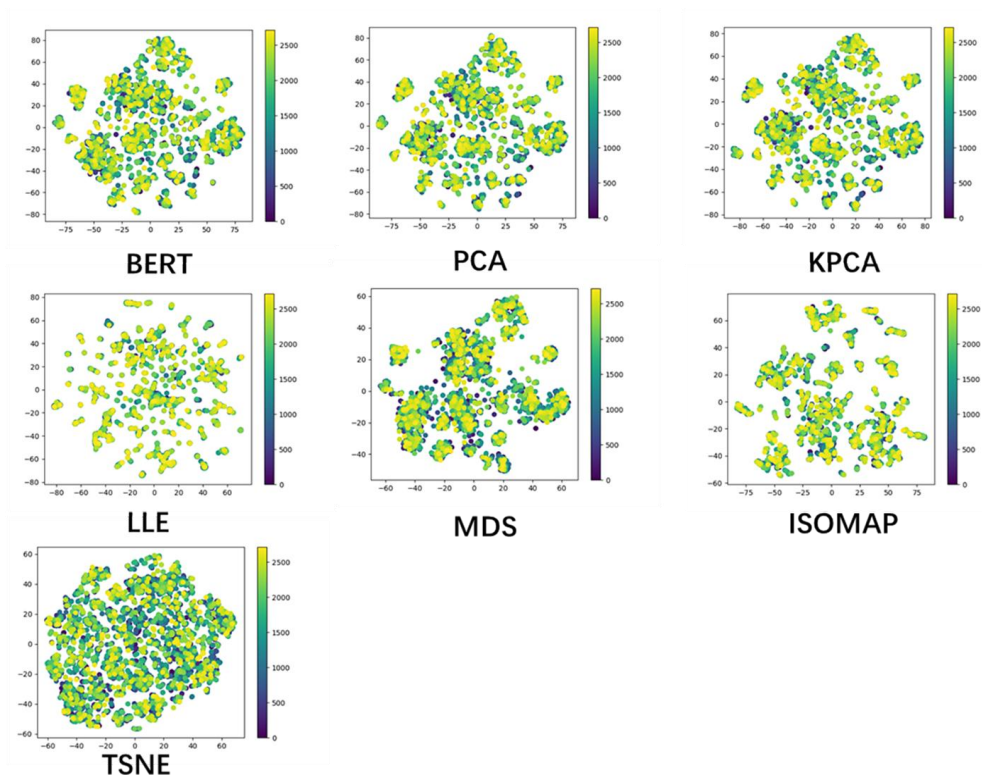

Fig.3 The 2D plane distribution of BERT and its reduced versions by different methods.

Table 1. The difference in self-similarity between the original BERT and reduced versions by using different methods.

| method | PCA | KPCA | LLE | MDS | ISOMAP | t-SNE |
| --- | --- | --- | --- | --- | --- | --- |
| gap | 0.1349 | 0.1350 | 0.1393 | 0.1375 | 0.2124 | 0.2599 |

The original 768-dimensional vectors encoded by BERT are compressed to 256, 128, and 64 dimensions using different linear layers, respectively. As shown in Fig. 4, the relationship between evaluation accuracy and training epochs for the equation formulation task is presented. To investigate the impact of vector dimensionality on downstream tasks related to MWPs, specifically the effect of input dimensions on the decoder's output, we freeze BERT's parameters to ensure the input vectors remain consistent. During training, the parameters of both the compressor and decoder are trainable to fully explore the potential of the input vectors.

It can be observed that as the dimensionality of the input vectors decreases, the final accuracy achieved after training also gradually decreases. For the study of vector compressibility, the critical factor is the accuracy relative to the pre-compression accuracy rather than the absolute accuracy. After compressing the dimensionality to one-third of the original (down to 256 dimensions), the accuracy shows only a slight decrease compared to before compression. Specifically, the accuracy drops from 27.76% to 21.04%, a decrease of 24.21%. Further compressing the vectors by half to 128 dimensions results in a significant decrease in accuracy, dropping from 21.04% to 12.93%, a reduction of 38.55%. This larger drop in accuracy indicates diminishing returns from further compression. Compressing further from 128 dimensions to 64 dimensions results in accuracy dropping from 12.93% to 7.52%, a decrease of 41.84%. This represents a smaller benefit from additional compression.

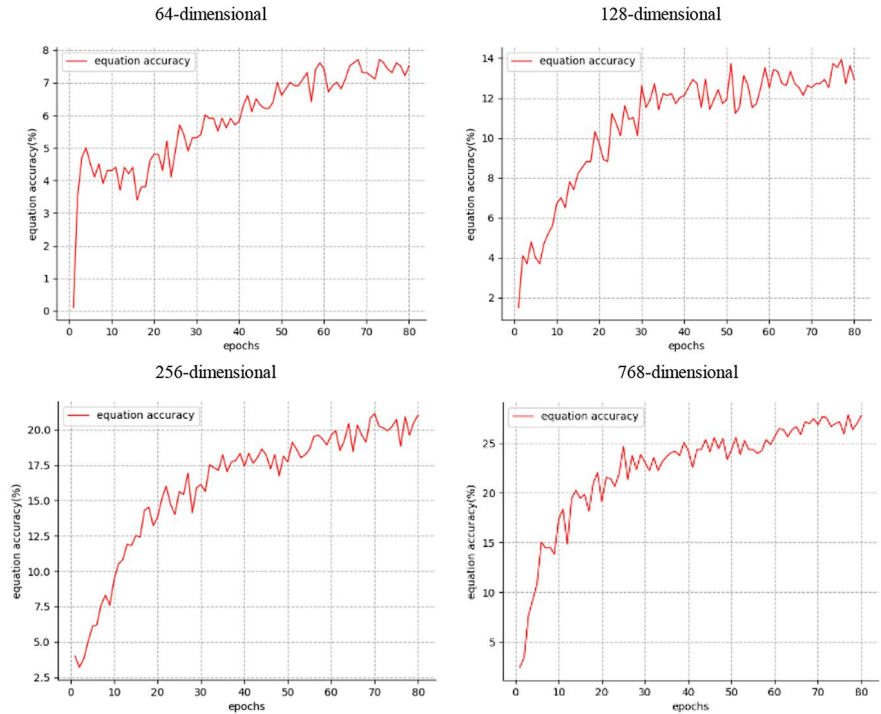

Fig.4 The relationship between evaluation accuracy and training epochs for the equation formulation task. The images in the top left, top right, bottom left, and bottom right correspond to encoded vector dimensions of 64, 128, 256, and 768, respectively.

As shown in Fig. 5, the relationship between evaluation accuracy and training epochs for the answer retrieval task is presented. It can be observed that as the dimensionality of the input vectors decreases, the final accuracy achieved after training also gradually decreases, following the same trend as observed in the equation formulation task. After compressing the dimensionality to one-third of the original (down to 256 dimensions), the accuracy shows only a slight decrease compared to before compression. Specifically, the accuracy drops from 33.27% to 23.75%, a decrease of 28.61%. Further compressing the vectors by half to 128 dimensions results in a significant decrease in accuracy, with the accuracy dropping from 23.75% to 14.33%, a reduction of 39.66%. Compressing further from 128 dimensions to 64 dimensions leads to accuracy dropping from 14.33% to 8.52%, a decrease of 40.54%.

The consistent pattern observed in both the answer retrieval and equation formulation tasks indicates that as the compression dimension gradually decreases, the rate of performance loss increases. The efficiency gains from compression will eventually be outweighed by the associated performance losses. The key challenge is to find an optimal compression dimension that achieves the best balance between performance and computational efficiency.

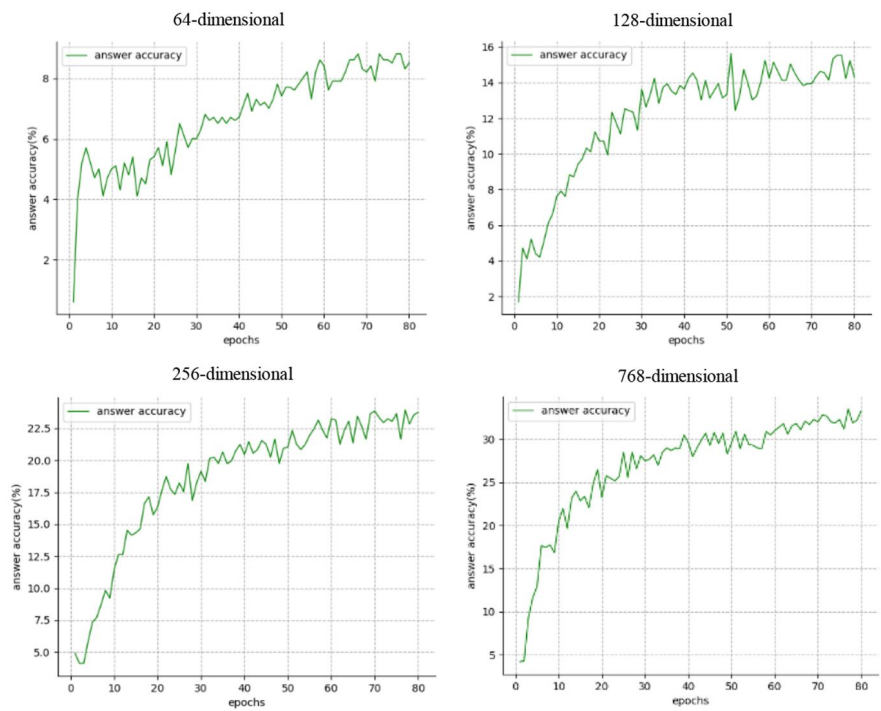

Fig.5 The relationship between evaluation accuracy and training epochs for the answer retrieval task. The images in the top left, top right, bottom left, and bottom right correspond to encoded vector dimensions of 64, 128, 256, and 768, respectively.

To validate the generalizability of this conclusion, we tested three different

compressors—PCA, LLE, and MDS—on the relation extraction task. The results are shown in Fig. 6. It is clear that PCA achieves higher accuracy after training compared to the other compressors across all three dimensions (256, 128, and 64). This suggests that the choice of compressor plays a crucial role in the final outcome. An effective compressor should retain as much useful information as possible while discarding redundant information, thereby significantly improving computational efficiency without sacrificing performance.

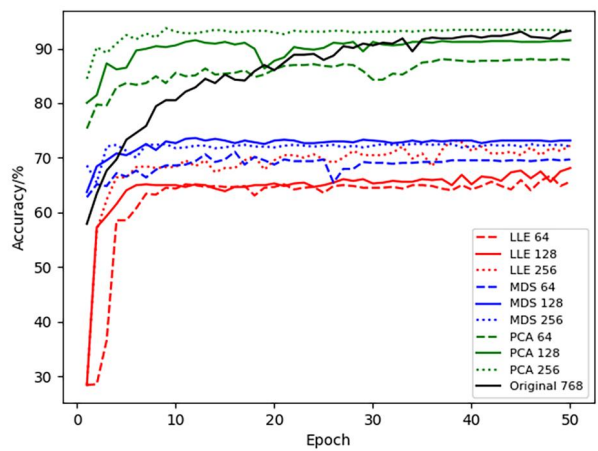

Fig.6 The relationship between evaluation accuracy and training epochs for the relation extraction task. The figure illustrates three compression methods: LLE, PCA, and MDS. Each method reduces the encoded vectors to three different dimensions—256, 128, and 64—resulting in a total of nine scenarios. Additionally, the uncompressed 768-dimensional vector is included in the figure for reference and comparison.

Compared to the original 768-dimensional vectors, which achieve a final accuracy of 93.25%, the compressed 256-dimensional vectors using PCA achieve a final accuracy of 93.41%, nearly identical, in the relation extraction task. This indicates that, as long as an appropriate compression method is selected, 256-dimensional encoded vectors can still effectively handle relation extraction tasks related to MWPs.

Considering all the results above, 256 dimensions emerge as an optimal choice, balancing computational efficiency and performance. This size, which is only one-third of the original encoding, can still maintain performance above 70% across all MWP-related tasks. Importantly, this conclusion is general and not specific to any particular decoder.

4.2 Knowledge Distillation from BERT

We chose 256 dimensions for vector encoding, as this dimensionality provides a balance between performance and computational efficiency, based on the conclusions from the previous experiments. As a result, all the hidden sizes of the transformer layers in the student model are set to 256. Following the approach of Jiao et al. [35], the distillation process is divided into two steps. A pre-trained BERT serves as the teacher model, while a randomly initialized model, as described in the Student Model section, is used as the base to obtain a preliminary student model after the first distillation step. In the second step, a fine-tuned BERT, trained on equation extraction and answer retrieval tasks, serves as the teacher model, and the preliminary student model is used as the base to obtain the final student model after distillation.

Table 2 compares the performance of the distilled model (after two distillation steps) and the undistilled model on the tasks of equation formulation and answer retrieval. On one hand, it is clear that the initial accuracy for equation formulation increases from 4.71% to 5.71%, marking an improvement of 21.23% after distillation. Similarly, the initial accuracy for answer retrieval increases from 5.21% to 7.31%, a more substantial improvement of 40.31% following distillation. These notable accuracy gains suggest that knowledge distillation plays a crucial role in improving performance for solving MWPs.

Table 2. Accuracy of the distilled and undistilled model for equation and answer tasks.

|  | Accuracy | undistilled | distilled | improvement |
|---|---|---|---|---|
| Initial | Equation Accuracy | 4.71% | 5.71% | 21.23% |
|  | Answer Accuracy | 5.21% | 7.31% | 40.31% |
| Final | Equation Accuracy | 26.95% | 27.25% | 1.11% |
|  | Answer Accuracy | 32.87% | 33.37% | 1.52% |

On the other hand, the final accuracy for equation formulation after training increases from 26.95% to 27.25%, a slight improvement of 1.11% following distillation. Similarly, the final accuracy for answer retrieval after training increases from 32.87% to 33.37%, a modest improvement of 1.52%. These minimal accuracy improvements suggest that the final training accuracy is more dependent on the model's network architecture. With sufficient training, identical network structures can achieve comparable performance, highlighting the importance of the underlying model architecture.

The final student model contains only 1/12 of the parameters of BERT-base while maintaining about 90% of the performance across both equation formulation and answer retrieval tasks. This demonstrates that the performance-to-size ratio improves after distillation. Furthermore, this distillation method does not rely on a specific decoder, but instead utilizes two encoders, illustrating its generalizability.

## 5. Discussion

We further investigate why the encoded vectors can be compressed for solving MWP,

aiming to achieve the optimal balance between performance and computational efficiency. Drawing inspiration from the syntax-semantics model proposed by Yu et al. [6], we hypothesize that understanding the parts-of-speech alone—without needing to grasp specific entities—may be a key factor.

5.1 Compressibility for Part-of-Speech Tagging Task

To validate our hypothesis, we first assigned the pre-trained BERT model the task of part-of-speech tagging after compressing the vectors to different dimensionalities. As shown in Fig. 7, although the initial evaluation accuracy slightly decreases with the reduction in vector dimensions, the evaluation accuracy after training remains nearly the same. This indicates that the compressed vectors still retain sufficient capability to perform the part-of-speech tagging task, suggesting that the original 768-dimensional vectors contain redundancy for this task. According to the syntax-semantics theory, this redundancy arises from the inclusion of specific entity information in natural language.

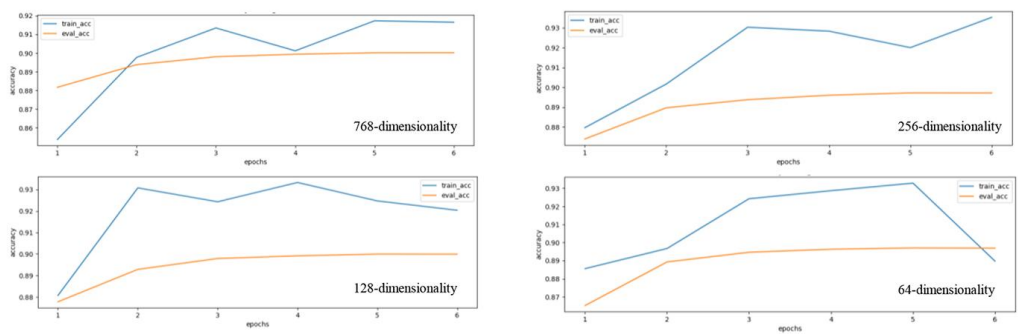

Fig.7 Training and evaluation accuracy for the part-of-speech tagging task. The images in the top left, top right, bottom left, and bottom right correspond to encoded vector dimensions of 768, 256, 128, and 64, respectively.

5.2 Relationship Between MWP and Part-of-Speech

The previous section discussed the compressibility of encoding vectors for the part-of-speech tagging task. To position this as one of the intrinsic reasons for the compressibility of vectors in MWP tasks, we must clarify the relationship between the two.

According to the syntax-semantics theory, the focus during template refinement is often on the syntactic structure of sentences and the part-of-speech tags of the words. Words that indicate logical and quantitative relations are more crucial than other words. Nouns, which represent specific entities, rarely indicate logical relations. In mathematical contexts, nouns often serve as placeholders, carrying minimal importance.

To validate this point, we design an experiment using the aforementioned TinyBERT model. We evaluate both a randomly initialized model and a model pre-trained on the part-of-speech tagging task, assessing their performance on formulating equations and obtaining answers. As shown in Fig. 8, both the initial accuracy and the final accuracy after training of the model pre-trained on the part-of-speech tagging task are significantly higher than the model without pre-training. This strongly suggests that part-of-speech tagging is highly beneficial for MWP-related tasks, aligning with the syntax-semantics theory.

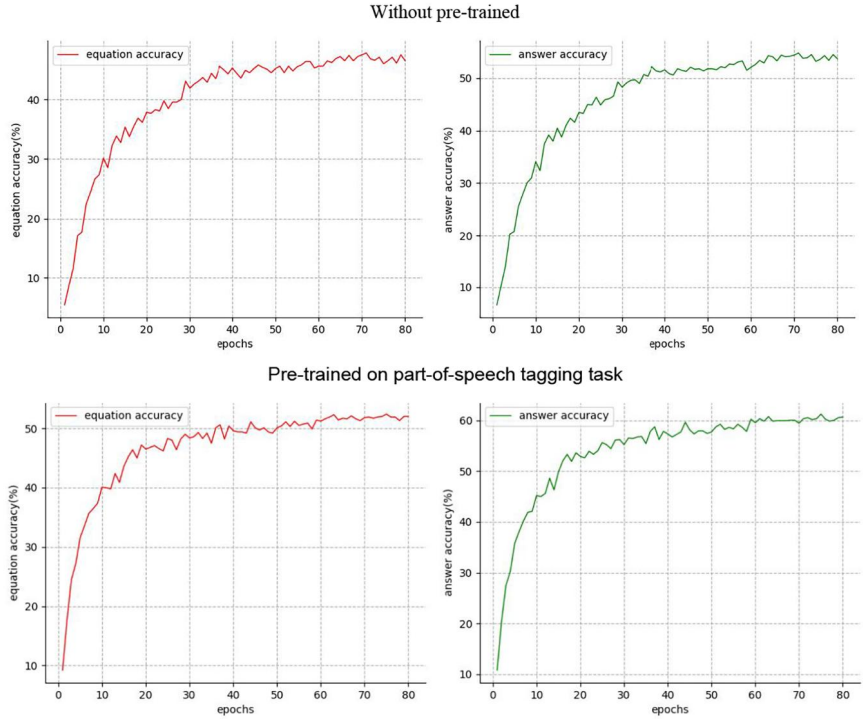

Fig.8 The relationship between evaluation accuracy and training epochs for both the equation formulation and answer retrieval tasks. The top two graphs represent models without part-of-speech tagging pre-training, while the bottom two graphs represent models with part-of-speech tagging pre-training.

Additionally, we examine the encoded vectors with the largest values in these dimensions and identify their corresponding tokens, as shown in Table 3. These tokens represent a significant proportion of the overall semantics of the sentence. It is clear that elements like punctuation, numbers, and quantity play crucial roles in problem understanding. This suggests that, in order to correctly interpret the semantics of mathematical problems, it is vital to focus on key phrases (such as "how much", "less than", "add", etc.) and specific parts-of-speech information. Therefore, when compressing vectors, it is essential not to lose this critical information.

Table 3. The key information for understanding the math problems.

| Sort | No.1 | No.2 | No.3 | No.4 | No.5 |
|---|---|---|---|---|---|
|  | punctuation | noun | numbers | Keywords | quantity |

## 6. Conclusion

This work explores the compressibility of encoded vectors for tasks related to MWP (Mathematical Word Problems). We employed four different vector dimensionality reduction methods: linear layer, PCA, LLE, t-SNE, ISOMAP, and MDS, to compress the vectors. Three decoders—focused on extracting relations, formulating equations, and obtaining answers—were used to evaluate the compressed vectors. While the effectiveness of compression varies depending on the method and the downstream task, all results suggest that the 768-dimensional vectors encoded by BERT contain redundancy for mathematical tasks. Therefore, compressing the vectors to an appropriate dimension can achieve an optimal balance between computational efficiency and performance.

A student model, consisting of an embedding layer and three transformer layers, was distilled from the BERT-base. The performance-to-size ratio improved after distillation. Notably, this distillation method involved only an encoder and no specific decoder, demonstrating a certain level of generalizability.

Additionally, we explored the underlying mechanism and found that part-of-speech information, rather than entity-specific details, plays a more crucial role in MWP. This may be a key reason behind the compressibility of the encoded vectors.

**Acknowledgments**

This work was financially supported by self-determined research funds of CCNU from the colleges' basic research and operation of MOE (Grant No. CCNU24XJ020) and National Natural Science Foundation of China under Grant 62177025.

**References**

[1] Floridi, L. and M. Chiriatti, GPT-3: Its nature, scope, limits, and consequences. Minds and Machines, 2020. 30: p. 681-694.

[2] Kulik, J.A. and J.D. Fletcher, Effectiveness of intelligent tutoring systems: a meta-analytic review. Review of educational research, 2016. 86(1): p. 42-78.

[3] Brown, T., et al., Language models are few-shot learners. Advances in neural information processing systems, 2020. 33: p. 1877-1901.

[4] Romera-Paredes, B., et al., Mathematical discoveries from program search with large language models. Nature, 2024. 625(7995): p. 468-475.

[5] Imani, S., L. Du and H. Shrivastava, Mathprompter: Mathematical reasoning using large language models. arXiv preprint arXiv:2303.05398, 2023.

[6] Yu, X., et al., Solving arithmetic word problems by synergizing syntax-semantics extractor for explicit relations and neural network miner for implicit relations. Complex & Intelligent Systems, 2023. 9(1): p. 697-717.

[7] Yao, J., Z. Zhou and Q. Wang. Solving math word problem with problem type classification. in CCF International Conference on Natural Language Processing and Chinese Computing. 2023: Springer.

[8] Zong, M. and B. Krishnamachari. Solving math word problems concerning systems of equations with gpt-3. in Proceedings of the AAAI Conference on Artificial Intelligence. 2023.

[9] Lin, X., et al. Hms: A hierarchical solver with dependency-enhanced understanding for math word problem. in Proceedings of the AAAI conference on artificial intelligence. 2021.

[10] Hong, Y., et al. Learning by fixing: Solving math word problems with weak supervision. in Proceedings of the AAAI conference on artificial intelligence. 2021.

[11] Yeo, S., J. Moon and D. Kim, Transforming mathematics education with AI: Innovations, implementations, and insights. The Mathematical Education, 2024. 63(2): p. 387-392.

[12] Blakeney, C., et al., Parallel blockwise knowledge distillation for deep neural network compression. IEEE Transactions on Parallel and Distributed Systems, 2020. 32(7): p. 1765-1776.

[13] Alzoubi, Y.I. and A. Mishra, Green artificial intelligence initiatives: Potentials and challenges. Journal of Cleaner Production, 2024: p. 143090.

[14] Bolón-Canedo, V., et al., A review of green artificial intelligence: Towards a more sustainable future. Neurocomputing, 2024: p. 128096.

[15] Irugalbandara, C., et al. Scaling Down to Scale Up: A Cost-Benefit Analysis of Replacing OpenAI's LLM with Open Source SLMs in Production. in 2024 IEEE International Symposium on Performance Analysis of Systems and Software (ISPASS). 2024: IEEE.

[16] Tambe, T., et al. Edgebert: Sentence-level energy optimizations for latency-aware multi-task nlp inference. in MICRO-54: 54th Annual IEEE/ACM International Symposium on Microarchitecture. 2021.

[17] Yu, X., H. Sun and C. Sun, A relation-centric algorithm for solving text-diagram function problems. Journal of King Saud University-Computer and Information Sciences, 2022. 34(10): p. 8972-8984.

[18] Lyu, X., X. Yu and R. Peng, Vector relation acquisition and scene knowledge for solving arithmetic word problems. Journal of King Saud University-Computer and Information Sciences, 2023. 35(8): p. 101673.

[19] Lyu, X. and X. Yu. Solving explicit arithmetic word problems via using vectorized syntax-semantics model. in 2021 IEEE International Conference on Engineering, Technology & Education (TALE). 2021: IEEE.

[20] He, B., et al., Comparative study of typical neural solvers in solving math word problems. Complex & Intelligent Systems, 2024: p. 1-26.

[21] Wang, Y., X. Liu and S. Shi. Deep neural solver for math word problems. in Proceedings of the 2017 conference on empirical methods in natural language processing. 2017.

[22] Zhang, D., et al., The gap of semantic parsing: A survey on automatic math word problem solvers. IEEE transactions on pattern analysis and machine intelligence, 2019. 42(9): p. 2287-2305.

[23] Wang, L., et al., Translating a math word problem to an expression tree. arXiv preprint arXiv:1811.05632, 2018.

[24] Liu, Q., et al. Tree-structured decoding for solving math word problems. in Proceedings of the 2019 conference on empirical methods in natural language processing and the 9th international joint conference on natural language processing (EMNLP-IJCNLP). 2019.

[25] Zhang, J., et al. Graph-to-tree learning for solving math word problems. in Proceedings of the 58th Annual Meeting of the Association for Computational Linguistics. 2020.

[26] Kenton, J.D.M.C. and L.K. Toutanova. BERT: Pre-training of Deep Bidirectional Transformers for Language Understanding. in Proceedings of NAACL-HLT. 2019.

[27] Radford, A., et al., Language models are unsupervised multitask learners. OpenAI blog, 2019. 1(8): p. 9.

[28] Liang, Z., et al. MWP-BERT: Numeracy-Augmented Pre-training for Math Word Problem Solving. in Findings of the Association for Computational Linguistics: NAACL 2022. 2022.

[29] Chowdhery, A., et al., Palm: Scaling language modeling with pathways. Journal of Machine Learning Research, 2023. 24(240): p. 1-113.

[30] Lewkowycz, A., et al., Solving quantitative reasoning problems with language models. Advances in Neural Information Processing Systems, 2022. 35: p. 3843-3857.

[31] Touvron, H., et al., Llama: Open and efficient foundation language models. arXiv preprint arXiv:2302.13971, 2023.

[32] Chen, J., et al., Skills-in-context prompting: Unlocking compositionality in large language models. arXiv preprint arXiv:2308.00304, 2023.

[33] Wei, J., et al., Chain-of-thought prompting elicits reasoning in large language models. Advances in neural information processing systems, 2022. 35: p. 24824-24837.

[34] Zhou, A., et al., Solving challenging math word problems using gpt-4 code interpreter with code-based self-verification. arXiv preprint arXiv:2308.07921, 2023.

[35] Jiao, X., et al., Tinybert: Distilling bert for natural language understanding. arXiv preprint arXiv:1909.10351, 2019.

[36] Hinton, G., O. Vinyals and J. Dean, Distilling the Knowledge in a Neural Network. stat, 2015. 1050: p. 9.

[37] Gou, J., et al., Knowledge distillation: A survey. International Journal of Computer Vision, 2021. 129(6): p. 1789-1819.

[38] Cho, J.H. and B. Hariharan. On the efficacy of knowledge distillation. in Proceedings of the IEEE/CVF international conference on computer vision. 2019.

[39] Tang, R., et al., Distilling task-specific knowledge from bert into simple neural networks. arXiv preprint arXiv:1903.12136, 2019.

[40] Sun, S., et al. Patient Knowledge Distillation for BERT Model Compression. in Proceedings of the 2019 Conference on Empirical Methods in Natural Language Processing and the 9th International Joint Conference on Natural Language Processing (EMNLP-IJCNLP). 2019.

[41] Sun, Z., et al. MobileBERT: a Compact Task-Agnostic BERT for Resource-Limited Devices. in Proceedings of the 58th Annual Meeting of the Association for Computational Linguistics. 2020.

[42]Wang, W., et al., Minilm: Deep self-attention distillation for task-agnostic compression of pre-trained transformers. Advances in Neural Information Processing Systems, 2020. 33: p. 5776-5788.

[43]Abdi, H. and L.J. Williams, Principal component analysis. Wiley interdisciplinary reviews: computational statistics, 2010. 2(4): p. 433-459.

[44]Roweis, S.T. and L.K. Saul, Nonlinear dimensionality reduction by locally linear embedding. science, 2000. 290(5500): p. 2323-2326.

[45]Borg, I. and P.J. Groenen, Modern multidimensional scaling: Theory and applications. 2007: Springer Science & Business Media.

[46]Van der Maaten L, Hinton G. Visualizing data using t-SNE[J]. Journal of machine learning research, 2008, 9(11).

[47]Tenenbaum J B, Silva V, Langford J C. A global geometric framework for nonlinear dimensionality reduction[J]. science, 2000, 290(5500): 2319-2323.